\documentclass[conference]{IEEEtran}
\IEEEoverridecommandlockouts
\usepackage{cite}
\usepackage{amsmath,amssymb,amsfonts}
\usepackage{algorithmic}
\usepackage{graphicx}
\usepackage{textcomp}
\usepackage{xcolor}
\graphicspath{{..figures/}{../figures/}}
\DeclareGraphicsExtensions{.pdf,.jpeg,.png}
\def\BibTeX{{\rm B\kern-.05em{\sc i\kern-.025em b}\kern-.08em
    T\kern-.1667em\lower.7ex\hbox{E}\kern-.125emX}}

\newcommand{\blockcomment}[1]{}

\begin{document}

\title{Stabilized Adaptive Steering for 3D Sonar Microphone Arrays with IMU Sensor Fusion}

\blockcomment{
\author{\IEEEauthorblockN{Wouter Jansen}
\IEEEauthorblockA{\textit{FTI Cosys-Lab, University of Antwerp}\\ Antwerp, Belgium \\
\textit{Flanders Make Strategic Research Centre}\\ Lommel, Belgium\\
wouter.jansen@uantwerpen.be}
\IEEEauthorblockN{Dennis Laurijssen}
\IEEEauthorblockA{\textit{FTI Cosys-Lab, University of Antwerp}\\ Antwerp, Belgium \\
\textit{Flanders Make Strategic Research Centre}\\ Lommel, Belgium\\
dennis.laurijssen@uantwerpen.be}
\and
\IEEEauthorblockN{Jan Steckel}
\IEEEauthorblockA{\textit{FTI Cosys-Lab, University of Antwerp}\\ Antwerp, Belgium \\
\textit{Flanders Make Strategic Research Centre}\\ Lommel, Belgium\\
jan.steckel@uantwerpen.be}
}
}

\author{
    \IEEEauthorblockN{
        Wouter Jansen\IEEEauthorrefmark{1}\IEEEauthorrefmark{2}\IEEEauthorrefmark{3},
        Dennis Laurijssen\IEEEauthorrefmark{1}\IEEEauthorrefmark{2},
        Jan Steckel\IEEEauthorrefmark{1}\IEEEauthorrefmark{2}
    }
    \IEEEauthorblockA{\IEEEauthorrefmark{1}Cosys-Lab, Faculty of Applied Engineering, University of Antwerp, Antwerp, Belgium}
    \IEEEauthorblockA{\IEEEauthorrefmark{2}Flanders Make Strategic Research Centre, Lommel, Belgium}
    \IEEEauthorblockA{\IEEEauthorrefmark{3}wouter.jansen@uantwerpen.be}
 }

\maketitle

\begin{abstract}
This paper presents a novel software-based approach to stabilizing the acoustic images for in-air 3D sonars. Due to uneven terrain, traditional static beamforming techniques can be misaligned, causing inaccurate measurements and imaging artifacts. Furthermore, mechanical stabilization can be more costly and prone to failure. We propose using an adaptive conventional beamforming approach by fusing it with real-time IMU data to adjust the sonar array's steering matrix dynamically based on the elevation tilt angle caused by the uneven ground. Additionally, we propose gaining compensation to offset emission energy loss due to the transducer's directivity pattern and validate our approach through various experiments, which show significant improvements in temporal consistency in the acoustic images. We implemented a GPU-accelerated software system that operates in real-time with an average execution time of 210ms, meeting autonomous navigation requirements.
\end{abstract}


\begin{figure*}
    \centering
    \includegraphics[width=1\linewidth]{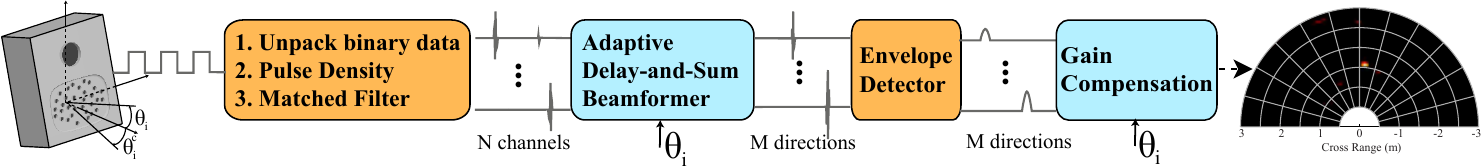}
    \caption{The processing pipeline from the raw sensor data of the 3D sonar sensor to acoustic images. The adaptive DAS-beamformer uses the measured tilt angle \(\theta_i\) captured by the IMU and gains compensation to stabilize the acoustic image result.}
    \label{fig:ertis_processing}
\end{figure*}

\section{Introduction}\label{sec:intro}
When observing the environment, humans are constantly adapting the gaze of our eyes and our head's rotation to what we are doing. For example, when driving a motor vehicle, we focus on the lane ahead, do far-distance scanning, look into our mirrors, etc. \cite{tawari_where_2014}. Similarly, when preparing food or doing housework, our eyes are focused on the used object. However, our eyes typically move on to the next object in a saccadic fashion in the preparation sequence before completing the preceding action \cite{land_what_2001}. Research into scene analysis by animals has provided researchers with much insight to develop algorithmic methods for robotic applications. Animals can selectively gather information, analyze cues, and adapt their sensing methodology accordingly\cite{lewicki_scene_2014}. Birds are a famous example of fully stabilizing their heads relative to their surroundings \cite{pete_role_2015}.
Another example is the emitted calls from echolocating bats, which can be highly directional in some species, especially in cluttered environments. These animals steer their emission beam through head motion. That way, they can enhance the information capture to suppress environmental noise and focus on their subject of interest, such as their insect prey \cite{thomas_echolocation_2002, lee_tongue-driven_2017, verreycken_bio-acoustic_2021}.

Ultrasonic in-air sonar sensors are bio-inspired in their design and utilization by these bat species. When using a dense microphone array such as developed by Steckel et al. \cite{kerstens_ertis_2019, laurijssen_hiris_2024, verellen_urtis_2020}, the signal processing often relies on beamforming techniques to steer the array into the directions of interest and to suppress the noise from others \cite{krim_two_1996}. In conventional beamforming, such as delay-and-sum (DAS) beamforming, the weights and parameters are fixed to provide a specific response, regardless of the input data. On the other hand, adaptive beamforming techniques are data-dependent. For example, the well-known MVDR algorithm determines the weights by maximizing the SNR over the array output. In general, when targeting real-time applications, conventional beamforming is sufficient. It has the benefit that its simplicity results in lower computational requirements and graceful degradation under high noise conditions.

Ultrasonic in-air sonar sensors have proven to work in autonomous navigation and mapping applications \cite{jansen_real-time_2022, steckel_biomimetic_2012}. In these applications, conventional DAS-beamforming was used to scan the horizontal plane in front of the sensor. In the experimental results, however, the ground on which the autonomous vehicle moved was always a perfectly flat surface. Suppose one were to introduce rough ground in, for example, an agricultural field or other outdoor environment, the fixed beamforming scan pattern would cause it to no longer align with the horizontal plane (the plane perpendicular to the gravity axis). This misalignment could cause incorrect sensor measurements and potentially negatively influence the underlying application. Furthermore, the single non-omnidirectional transducer used in these experiments, emitting a broadband chirp call, will no longer be focused directly forward and, due to its directivity pattern, cause a difference in reflected energy at the directions of interest. 

The microphone array and emitter are typically mounted in fixed positions on the vehicle and cannot be dynamically moved to adjust for uneven ground. This is in contrast to what is often done with other sensor modalities that require stabilization, such as cameras that can use external gimbals \cite{rajesh_camera_2015} or that have built-in Optical Image Stabilization (OIS) for achieving nearly perfect stabilized images. While an external gimbal could be applied to a sonar sensor \cite{schillebeeckx_bio-inspired_2008}, it would introduce additional complexity, costs, points of failure, weight, and occupy space. Next to hardware-based stabilization, cameras often also rely on software solutions such as Electronic Image Stabilization (EIS). This software-based method uses an Inertial Measurement Unit (IMU), which includes accelerometers and gyroscopes, to detect motion. The software then adjusts the image digitally to compensate for the detected movement. This is often done by cropping and shifting frames in a video to smooth out motion. 

In this paper, we introduce electronic acoustic image stabilization by fusing data from an IMU sensor and using it to compensate for the tilt offset caused by the uneven ground. This is done by adapting the DAS-beamforming steering matrix in real-time. Furthermore, the transducer's emission energy loss is offset by a factor calculated during a calibration measurement to account for the transducer emission characteristic. We perform a set of validation and real-world scenario experiments to prove this technique and demonstrate how it operates in real-time. This cost-effective and versatile solution shows minimal additional latency, all while significantly increasing accuracy compared to unstabilized acoustic images. 

\begin{figure}[b]
    \centering
    \includegraphics[width=1\linewidth]{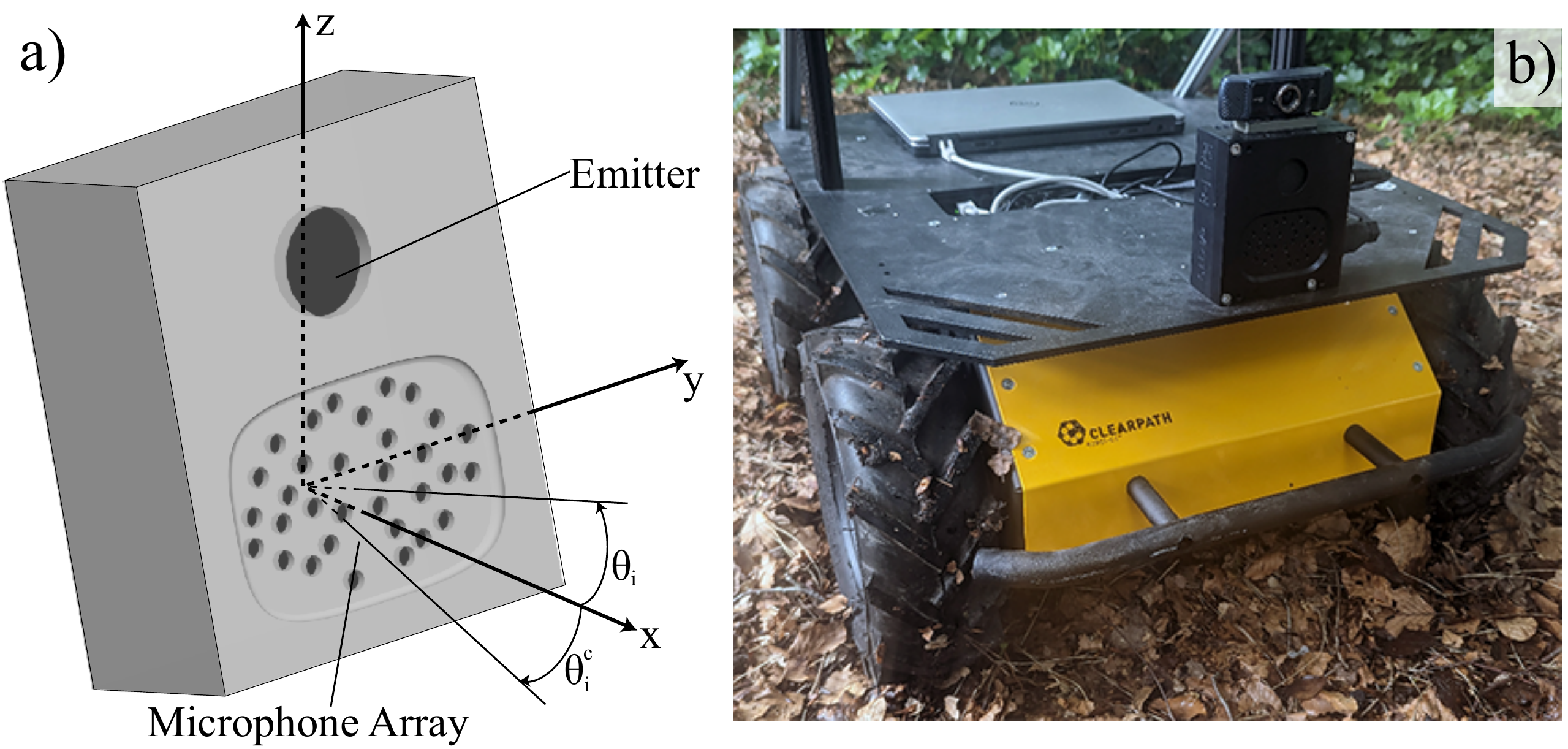}
    \caption{The embedded Real-Time Imaging Sonar (eRTIS) (a) is shown as a drawing at a tilted elevation angle \(\theta_i\). The opposite angle \(\theta_{i}^c=-\theta_i\) needs to be used to stabilize this sensor. (b) shows the eRTIS sensor mounted on a Clearpath Husky ruggedized mobile platform for outdoor experiments.}
    \label{fig:ertis_coordinates_husky}
\end{figure}

\section{3D Sonar \& IMU Hardware} \label{sec:fusion}
The in-air sonar sensor utilized in this study is the embedded Real-Time Imaging Sonar (eRTIS). A comprehensive description of the sonar can be found in \cite{kerstens_ertis_2019,steckel_broadband_2013}. The latest iteration of this sensor incorporates an embedded NVIDIA Jetson platform, which provides GPU-accelerated signal processing algorithms to operate in real-time \cite{jansen_-air_2020}. This sensor can be seen in Figure \ref{fig:ertis_coordinates_husky}b. Additionally, a TDK InvenSense ICM-20948 IMU sensor is integrated within the device \cite{tdk_icm-20948_nodate}. A Python script was developed to sample this IMU sensor over I2C on the NVIDIA Jetson, utilizing the Attitude and Heading Reference System (AHRS) algorithm by Madgwick S. \cite{madgwick_ahrs_2014} to fuse the gyroscope, accelerometer, and magnetometer data from the IMU into a single measurement of elevation \(\theta_i\), as illustrated in Figure \ref{fig:ertis_coordinates_husky}a.

\section{Adaptive Steering \& Gain Compensation} \label{sec:algorithm}
The adaptive steering aims to stabilize the scanning pattern, ensuring it always performs a horizontal 2D scan, whether the surface is uneven or not. This involves compensating the angle \(\theta_i\) measured by the IMU to select the steering matrix for DAS-beamforming, which scans in directions with an elevation angle \(\theta_{i}^c=-\theta_i\). The 3D-sonar employs a GPU-accelerated CUDA implementation of the signal processing pipeline \cite{jansen_-air_2020}. Part of this algorithm includes reducing latency by pre-buffering all memory required for both static and dynamic variables, which encompasses the delay matrix needed for DAS-beamforming. To enable real-time adaptive steering, a predetermined range of angles \(\theta^c\) is established. In this study, this range was set between \(-30^\circ\) and \(30^\circ\) with an angular resolution of \(1^\circ\). The DAS-beamforming delay matrix was pre-generated and pre-loaded into GPU memory along with all variables of the DSP pipeline. The available GPU memory determines the angular resolution that can be utilized for adaptive steering. When the IMU reports a new angle \(-\theta_i\), the algorithm selects the closest matching angle within the predetermined range and uses the associated delay matrix.

This study's second novel aspect is a dynamic gain compensation applied to the resulting acoustic image to counteract the limited directivity pattern of the transducer emitting the chirp. The gain factors are calculated during calibration. The 3D sonar sensor is mounted on a pan-tilt unit and directed at a flat surface reflector at a predetermined range. The pan-tilt unit is adjusted to predetermined angles corresponding to the range used for adaptive steering, and a sonar measurement is taken. The energy amplitude and the tilt angle are recorded at the reflector's range. The entire set of energy amplitudes is normalized against the value at \(0^\circ\) elevation. This value is then used to multiply the acoustic image in the final step, compensating for the energy loss due to elevation changes caused by uneven surfaces. The entire processing pipeline employed in this study is depicted in the diagram in Figure \ref{fig:ertis_processing}.

\begin{figure*}
    \centering
    \includegraphics[width=1\linewidth]{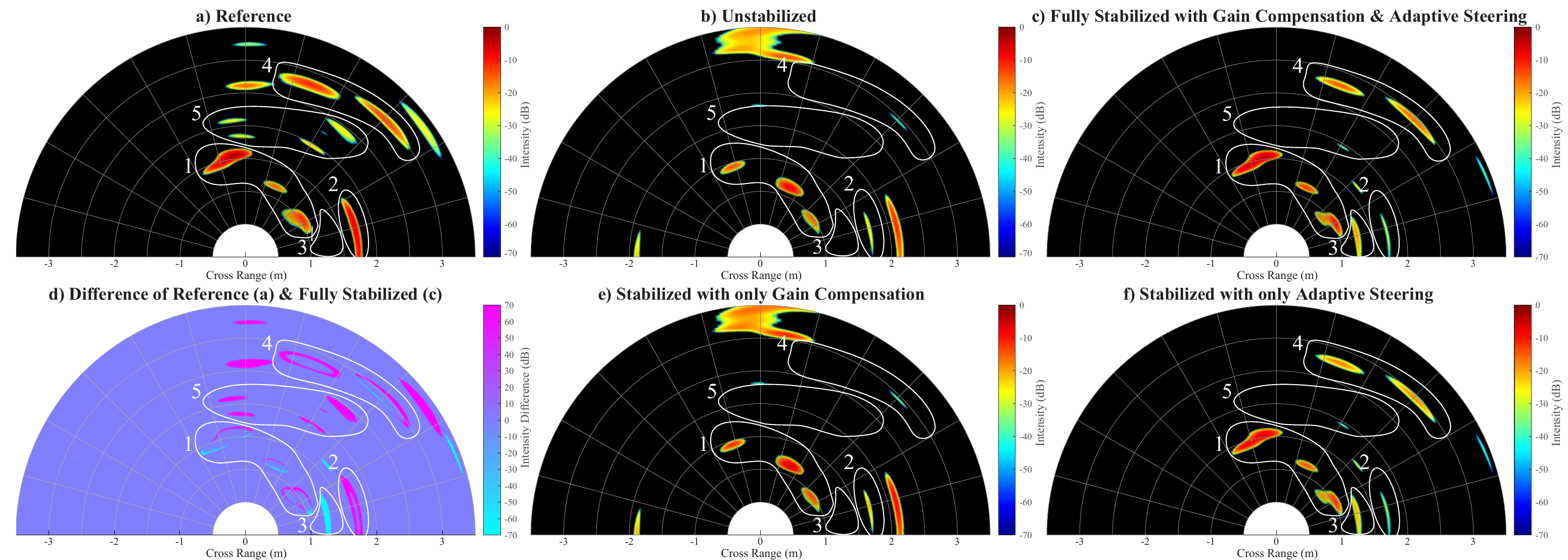}
    \caption{A single validation measurement with (a) the reference acoustic image at \(0^\circ\) elevation compared against a measurement at a tilt angle of \(\theta_i=-10.78^\circ\) shown in other images, with numbered reflector groups. (b) Unstabilized shows the loss of reflector groups 3 and 4. (c) Fully stabilized image at an adapted angle \(\theta_{i}^c=11^\circ\) with gain compensation (x1.34) restores these reflectors. Groups 1 and 2 appear in both. Group 5 are secondary reflections on the sensor, reflected away at a tilted angle. (d) Difference between reference and fully stabilized images. (e) and (f) show the two techniques separately.}
    \label{fig:still_frame}
\end{figure*}

\begin{figure}[!b]
    \centering
    \includegraphics[width=1\linewidth]{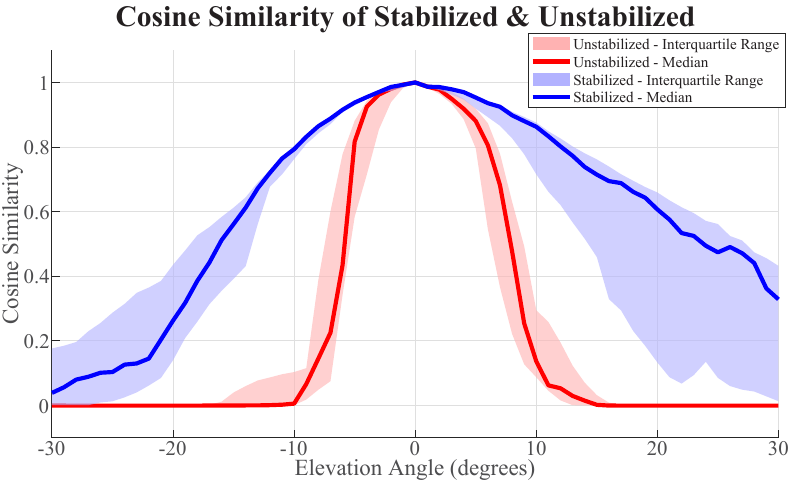}
    \caption{The validation results of all controlled experiments. Cosine similarity was used as a metric for image comparison between the (un)stabilized (at a certain tilt elevation angle) and reference acoustic images.}
    \label{fig:cosine_similarity_results}
\end{figure}

\section{Experimental Results}\label{sec:results}
We recorded ten datasets with the 3D sonar within indoor cluttered environments on a pan-tilt unit that controlled the elevation angle to validate the stabilization techniques of adaptive steering and gain compensation. These tilted measurements were compared against the reference measurement \(0^\circ\) elevation. Testing was conducted at the same angles used for the adaptive steering, setting the pan-tilt unit between \(-30^\circ\) and \(30^\circ\) with an angular resolution of \(1^\circ\). 
\\A comparison and analysis of the stabilized and unstabilized acoustic images are presented in Figure \ref{fig:still_frame}. 

For statistical analysis, the cosine similarity was calculated between the (un)stabilized and reference acoustic images. The results, shown in Figure \ref{fig:cosine_similarity_results}, indicate that the stabilization offers limited benefit for small tilt angles. However, from \(\pm3^\circ\) to \(\pm10^\circ\), the stabilization provides a significant benefit as similarity remains high to the reference measurement, and the difference becomes more pronounced against the unstabilized. Above \(\pm10^\circ\) of tilt, it indicates that the benefit of these stabilization techniques plateaus and gradually declines. The results indicate poorer performance for negative elevation angles, likely because of ground reflections. 

In addition to the validation, several real-time datasets were recorded on rough terrain with a ruggedized mobile platform (Clearpath Husky), as seen in Figure \ref{fig:ertis_coordinates_husky}b. \\ \\While no ground-truth validation is possible because of the nature of the experimental setup, the data shows clear visual improvements in the acoustic images when using the stabilization algorithm. The results can be viewed on YouTube\cite{jansen_steadirtis_2024}. From these experiments, we could also calculate that the elevation tilt has a 95\% confidence interval between \(\--3.14^\circ\) to \(3.47^\circ\) with an absolute minimum of \(\--12.25^\circ\) and absolute maximum of \(30.64^\circ\), respectively. Within this confidence interval, the validation shows an effective stabilization result. 

Finally, aiming to achieve real-time stabilization, we calculated an average execution time of 210ms with a standard deviation of 5ms, meeting the target measurement frequency of the existing autonomous applications that these sensors are used with \cite{jansen_real-time_2022, steckel_biomimetic_2012}.

\section{Conclusions \& Future Work}\label{sec:conclusions}
In this paper, we proposed a sensor fusion approach between in-air 3D Sonar and IMU to stabilize the acoustic images from horizontal tilt mismatches on mobile ground vehicles. We present two novel techniques for this stabilization: on the one hand, compensating for the gain loss because of the transducer's narrow directivity pattern, and on the other hand, adapting the DAS-beamforming steering matrix to direct forward consistently. The experimental validation shows that the stabilization is effective within the tilt range expected for uneven ground in mobile robots used for navigation and mapping in indoor and outdoor environments. 

Further experimentation within these applications is necessary to establish the practical utility of this stabilization technique. Due to the GPU-accelerated software implementation with pre-loaded memory of all DAS-beamforming steering matrices, the stabilization operates in real-time at nearly 5Hz, meeting existing autonomous navigation requirements. Future work could address rolling noise caused by ground irregularities and attempt to stabilize 3D acoustic point clouds.  

\newpage
\bibliographystyle{IEEEtran}
\bibliography{main.bib}

\end{document}